\documentclass{article}

\usepackage{microtype}
\usepackage{graphicx}
\usepackage{subfigure}
\usepackage{booktabs} 

\usepackage{times}
\usepackage{epsfig}
\usepackage{graphicx}
\usepackage{amsmath}
\usepackage{amssymb}

\usepackage{bigstrut}
\usepackage{bm}
\usepackage{wrapfig}
\usepackage{float}
\usepackage{hyperref}

\newcommand{\cD}{\mathcal{D}}

\newcommand{\cA}{\mathcal{A}}
\newcommand{\cL}{\mathcal{L}}
\newcommand{\cH}{\mathcal{H}}
\newcommand{\cY}{\mathcal{Y}}

\usepackage[arxiv]{icml2021}

\icmltitlerunning{Meta-Continual Generalized Zero-Shot Learning}

\begin{document}

\twocolumn[
\icmltitle{Meta-Learned Attribute Self-Gating\\for Continual Generalized Zero-Shot Learning}



\icmlsetsymbol{equal}{*}

\begin{icmlauthorlist}
\icmlauthor{Vinay Kumar Verma}{to}
\icmlauthor{Kevin Liang}{to}
\icmlauthor{Nikhil Mehta}{to}
\icmlauthor{Lawrence Carin}{to}
\end{icmlauthorlist}

\icmlaffiliation{to}{Duke University, NC, USA}

\icmlcorrespondingauthor{Vinay Verma}{vinaykumar.verma@duke.edu }

\icmlkeywords{Continual Learning, Zero-Shot Learning, Meta-Learning}

\vskip 0.3in
]



\printAffiliationsAndNotice{}  

\begin{abstract}
Zero-shot learning (ZSL) has been shown to be a promising approach to generalizing a model to categories unseen during training by leveraging class attributes, but challenges still remain.
Recently, methods using generative models to combat bias towards classes seen during training have pushed the state of the art of ZSL, but these generative models can be slow or computationally expensive to train. Additionally, while many previous ZSL methods assume a one-time adaptation to unseen classes, in reality, the world is always changing, necessitating a constant adjustment for deployed models. Models unprepared to handle a sequential stream of data are likely to experience catastrophic forgetting. We propose a meta-continual zero-shot learning (MCZSL) approach to address both these issues. In particular, by pairing self-gating of attributes and scaled class normalization with meta-learning based training, we are able to outperform state-of-the-art results while being able to train our models substantially faster ($>100\times$) than expensive generative-based approaches. We demonstrate this by performing experiments on five standard ZSL datasets (CUB, aPY, AWA1, AWA2 and SUN) in both generalized zero-shot learning and generalized continual zero-shot learning settings.
\end{abstract}

\section{Introduction}
Deep learning has demonstrated the ability to learn powerful models \cite{krizhevsky2012imagenet,he2016deep} given a sufficiently large, labeled dataset of a pre-defined set of classes.
However, such models often generalize poorly when asked to classify previously unseen classes that were not encountered during training.
In an ever-evolving world, in which new concepts or applications are to be expected, this brittleness can be an undesirable characteristic. 
While one could collect a new dataset and retrain a model, the associated time and costs make this rather inefficient.

In recent years, zero-shot learning (ZSL) \cite{akata2013label,norouzi2013zero,mishra2017generative,verma2019meta,skorokhodov2021class} has been proposed as an alternative framework.
Rather than having to collect more data and relearn the network upon encountering a previously unseen class, zero-shot approaches seek to leverage auxiliary information about these new classes, often in the form of class attributes.
This side information allows for reasoning about the relations between classes, enabling adaptation of the model to recognize samples from one of the novel classes.
In the more general setting, a ZSL model should be capable of classifying inputs from both seen and unseen classes; this more difficult setting \cite{xian2018zero,xian2018feature,vermageneralized} is commonly referred to as generalized zero-shot learning (GZSL).

Some of the strongest results in GZSL~\cite{felix2018multi,vermageneralized,schonfeld2019generalized,f-VAEGAN-D2} have come from approaches utilizing generative models. 
These approaches typically learn a generative mapping between the attributes and the data and then conditionally generate synthetic samples from the unseen class attributes.
The model can then be learned in the usual supervised manner on the joint dataset containing the seen class and generated unseen class data, mitigating model bias toward the seen classes.
While effective, training the requisite generative model, generating synthetic data, and training the model on this combined data can be an expensive process~\cite{narayan2020latent,li2019rethinking,verma2019meta} and requires all attributes of the unseen classes to be known \textit{a priori}, limiting the flexibility of such approaches. 

\begin{figure*}
	\label{fig:proposed_arch}
	\centering	
	\includegraphics[scale=0.8]{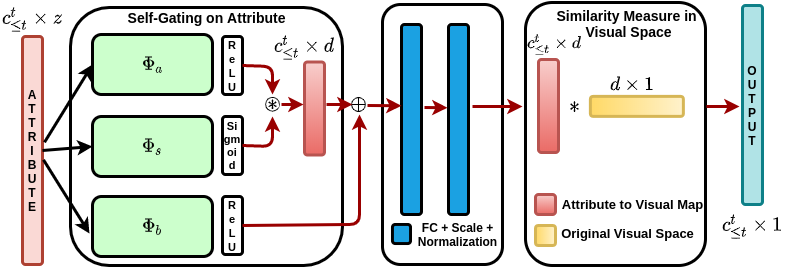}
	\caption{The proposed meta-continual zero-shot learning (MCZSL) model, with self-gating on attributes and scaled layer normalization.}
\end{figure*}

If assuming a one-time adaptation from a pre-determined set of training classes, the cost of such a one-off process may be considered acceptable in some circumstances, but it may present challenges if it must be repeated.
While ZSL methods commonly consider only one adaptation, in reality environments are often dynamic, and new class data may appear sequentially.
For example, if it is important for a model to be able to classify a previously unseen class, it is natural that a future data collection effort may later make labeled data from these classes available~\cite{skorokhodov2021class}.
Alternatively, changing requirements may require the model to learn from and then generalize to entirely new seen and unseen classes~\cite{gautam2020generalized}. 
In such cases, the model should be able to learn from new datasets without catastrophically forgetting~\cite{mccloskey1989catastrophic} previously seen data, even if that older data is no longer available in its entirety.
Thus, it is important that ZSL methods can work in continual learning settings as well. 

To address these issues, we propose meta-continual zero-shot learning (MCZSL).
We propose a novel self-gating on the attributes and scaled layer normalization that obviate the need for expensive generative models, resulting in a $100\times$ speed-up in training time.
We also show that training this model with a first-order meta-learning algorithm~\cite{nichol2018first} and a small reservoir of samples enables learning in a continual fashion, largely preventing catastrophic forgetting.
Experiments on CUB~\cite{CUB}, aPY~\cite{aPY}, AWA1~\cite{AWA1}, AWA2~\cite{xian2018zero} and SUN~\cite{patterson2012sun} demonstrate that MCZSL achieves state-of-the-art results in both GZSL and continual GZSL settings.

\section{Methods}
The proposed approach can be divide into three major components: ($i$) self-gating on the attribute, which helps discard noisy attribute dimensions, providing canonical, robust, class-specific information; ($ii$) normalization and scale, which can mitigate the seen class bias and play a key role in zero-shot learning \cite{skorokhodov2021class}; and ($iii$) meta-learning for few-shot learning \cite{nichol2018first}, which enables the model to learn a robust mapping when only a few samples are present. We describe here notation and each component in detail.

\subsection{Background and Notation}
Let $T_t=\{\cD_{tr}^t,\cD_{ts}^t\}$ be a task arriving at time $t$, where $\cD_{tr}^t$ and $\cD_{ts}^t$ are the train and test sets associated with the $t^{\mathrm{th}}$ task, respectively. In a continual learning setup, we assume that a set of tasks arrive sequentially, such that the training data for only the current task is made available. Let this sequence of tasks be $\{T_1,T_2,\dots T_K\}$, where at time $t$, the training data for only the $t^{\mathrm{th}}$ task is available. In continual learning, the goal is to learn a new task while preventing catastrophic forgetting of any previous tasks, therefore, at test the model is evaluated on the current and previous tasks that the model has encountered before. 

We assume that for a given task $t$, we have $c^s_t$ number of seen classes that we can use for training and $c^u_t$ unseen/novel classes that we are interested in adapting our model to. Analogous to GZSL, during test time we assume that samples come from any of the seen or unseen classes of the current or previous tasks, $i.e.$, the test samples for task $T_t$ contain $c_{\leq t}^s$ ($c_{\leq t}^s=\sum_{k=1}^tc_{k}^s$) number of seen classes and $c_{t}^u$ number of unseen classes, where $c_{t}^u$ will depend on the chosen evaluation protocol (see Sections \ref{sec:setting-1}-\ref{sec:setting2}). Assume $\cD_{tr}^t=\{x_i^t,y_i^t,a_{y_i},t\}_{i=1}^{N_{t}^s}$ to be the training data, where
$x_i^t\in \mathbb{R}^d$ is the visual feature of sample $i$ of task $t$, $y_i$ is the label of $x_i^t$, $a_{y_i}$ is the attribute/description vector of label $y_i$, $t$ is the task identifier (id) and $\cY_c$ is the label set for $c$ classes. Let $\cD_{ts}^t=\{\{x_i^t\}_{i=1}^{N_{t}^u},y_{x_i^t},\cA_t: \cA_t=\{a_y\}_{y=1}^{y=c_{\leq t}^s\cup c_{t}^u}\}$ be the testing data for the task $t$ and $y_{x_i^t}\in c_{\leq t}^s\cup c_{t}^u$. For each task $T_t$, the seen and unseen classes are disjoint, $i.e.$,  $\cY_{c_{t}^s}\cap \cY_{c_{t}^u}=\phi$. In each task $t$, seen and unseen classes have an associated attribute/description vector $a_y\in \cA_t$ that helps to transfer knowledge from seen to unseen classes. We use $N_{t}^s$ and $N_{t}^u$ to denote the number of train and test samples for the task $t$, respectively. We assume that there are a total of $S$ seen classes and $U$ unseen classes for each dataset, and the attribute set is $\cA: \cA\in\mathbb{R}^{S\cup U\times z}$, where $z$ is the attribute dimension. In the continual learning setup, test samples contain both unseen and seen classes up to task $t$. Therefore, the objective in continual ZSL is not only to predict the future task classes, but also to overcome catastrophic forgetting of the previous tasks.

\subsection{Self-Gating on Attributes}
\label{sec:selfattention}
Class attribute descriptions are a key component for positive forward transfer in ZSL. Models rely on the class description in order to successfully transfer knowledge from the seen classes to the unseen classes. These attributes are often either learned or manually annotated by humans, encapsulating all aspects of each class with a single set of attributes. Depending on the setting, intra-class variability can be significant; for example, members of a dog class can vary from a teacup poodle to Great Danes.
As a result, setting global attributes that are representative of the entire class can be a significant challenge. Because of the diversity of samples within a class, certain dimensions of an attribute may contain noisy or redundant information, which can degrade model performance. 

To address this, we develop a self-gating module for the attribute vector, that can continually learn from sequential data and provide a robust, class-representative vector satisfying most of the samples within the class. Let $\cA_{\leq t}^s$ be the seen class attribute up to the training of task $T_t$, where  $\cA_{\leq t}^s\in \mathbb{R}^{c_{\leq t}^s\times D}$. We introduce $\Phi_a$, $\Phi_s$, and $\Phi_b$ as three neural networks with different activation functions at the output layer. The self-gating on the attribute can be defined as: 
\begin{align}
\cA^s_{a_t^+} \enspace = \enspace &\mathrm{ReLU}\left(\Phi_a\left(\cA_{\leq t}^s\right)\right)*\sigma\left(\Phi_s\left(\cA_{\leq t}^s\right)\right) \nonumber \\ 
&+ \mathrm{ReLU}\left(\Phi_b\left(\cA_{\leq t}^s\right)\right)
\end{align}
where $\sigma$ is the sigmoid activation function. The function $\Phi_s$ with a sigmoid activation maps the projected attribute to the range $[0,1]$, giving weights to each dimension of the function $\Phi_a$. 
Values closer to one signify higher importance of a particular attribute dimension, while values closer to zero imply the opposite.
Finally, function $\Phi_b$ projects the attribute to the same space as $\Phi_a$ and $\Phi_b$, resulting in something similar to a bias vector, but learned through a different function. We empirically observe that $\Phi_a$ and $\Phi_b$ help learn a robust and global attribute vector while $\Phi_b$ helps stabilize model training. An overview of the self-gating block is illustrated in Figure~\ref{fig:proposed_arch}.

\subsection{Scaled Layer Normalization}
\label{sec:norm_scale}
Normalization techniques have become critical components of deep learning models. 
Popular normalizations like BatchNorm~\cite{ioffe2015batch}, GroupNorm~\cite{wu2018group}, and LayerNorm~\cite{ba2016layer} not only help stabilize and accelerate training of deep learning models but can also lead to significant improvements in final performance.
Recently, \citet{skorokhodov2021class} investigated the impact of class-wise layer normalization (CLN) on zero-shot learning. CLN was shown to result in significant improvement in ZSL models while also helping to overcome model bias towards the seen classes. The proposed scaled layer normalization (SLN) is motivated by CLN and scales the mean and variance vector by learnable parameters $\alpha$ and $\beta$. With $h_c\in \mathbb{R}^{d}$ being the output of the hidden layer of class $c$, we define SLN on $h_c$ as:
\begin{equation}
\label{eq:scaleCN}
SCN(h_c)=\frac{(h_c-\alpha\mu)}{\beta\sigma}
\end{equation} 
where $\{\alpha,\beta\}\in\mathbb{R}$ and $\{\mu,\sigma\}\in \mathbb{R}^d$. 
We apply SCN to each hidden layer, empirically observing that the proposed normalization leads to significant improvement.

\subsection{Reservoir Sampling} 
In continual learning settings, we assume that data arrive sequentially task by task, with only samples of the present task available in their entirety; samples of classes from previous tasks ($T_1,\dots, T_{t-1}$) are not accessible. Due to the tendency of neural networks to experience catastrophic forgetting~\cite{mccloskey1989catastrophic}, the model is likely to forget previously learned knowledge while learning new tasks.
While attribute self-gating and scaled layer normalization are highly effective for ZSL, by helping generalization to future classes, alone they do not prevent forgetting.

To mitigate catastrophic forgetting, our proposed model incorporates reservoir sampling~\cite{vitter1985random}, using a small and constant memory size to store selected samples from previous tasks and replaying these samples while training the current task. This has several advantages: ($i$) each task can be trained with constant time and computational resources, and ($ii$) the number of samples do not grow as the tasks increase.
Replaying samples from the reservoir effectively mitigates forgetting.
For a simple reservoir, each sample is selected for replay with probability $M/N$, where $M$ is the memory budget and $N$ is the number of samples selected so far.
We also tried using a ring buffer~\cite{lopez2017gradient}, but empirically found this to provide similar performance as the simple reservoir we employ.

\subsection{Training with Meta-Learning}
\label{sec:meta}
While reservoir sampling is effective for mitigating forgetting, it also has several drawbacks. As the task number grows, a constant memory budget means the number of samples for of each class diminishes, as the same amount of memory has to accommodate a large number of classes. Similarly, the current task will have a sizably larger number of samples for each class in the current task than the number of samples in the reservoir for classes of past tasks. Therefore, training can be difficult, resulting in models that generalize unevenly across all tasks/classes, as the model's learning is biased towards high-frequency classes. This issue is similar to the problem of few-shot learning, where we have to learn a model using only a few training examples present in the reservoir from the previous tasks.

To handle the aforementioned problems, we propose a meta-learning framework to train the model architecture. Meta-learning strategies have shown promising results for few-shot learning~\cite{finn2017model,nichol2018first}. Some of these approaches do this by training the model to learn a better initialization, from which it can quickly adapt to novel tasks with only a few samples; as such, we view this to naturally complement reservoir sampling. Our approach leverages Reptile~\cite{nichol2018first}, a relatively simple meta-learning approach using first-order gradient information, without storing gradients in memory for the inner loop. Let $f$ be our proposed model with parameters $\theta$, and $\tau$ be a batch of data in a $N$-way and $K$-shot setting (here we refer to batches instead of tasks to avoid confusion with tasks in a continual learning sense). Suppose $\cL_t$ is the loss over the batch $\tau$, given as:
\begin{equation}
\label{eq:loss}
\cL_\tau=\mathbb{E}_{(x,y,a)\sim\tau}l(f_\theta(a,x),y)
\end{equation}
where $l$ may be any suitable loss; in our experiments, we use cross-entropy loss. Assume that $U_\tau^z$ is an operator representing $z$ number of gradient update steps of the model parameter $\theta$, on the batch of data $\tau$. After $z$ gradient updates of the loss $\cL_\tau$, the model's new gradient is defined as:
\begin{equation}
\widetilde{\theta}=U_\tau^z(\theta)
\end{equation}
After the final gradient update, instead of updating the gradient in the direction of $(\widetilde{\theta}-\theta)$, Reptile considers $(\theta-\widetilde{\theta})$ itself as a gradient, resulting in a final update:
\begin{equation}
\label{eq:meta_finalupdate}
\theta\leftarrow\theta-\eta(\theta-\widetilde{\theta})
\end{equation}
which has been shown by \citet{nichol2018first} to approximate model-agnostic meta-learning~\cite{finn2017model}.
  
\vspace{-2mm}
\section{Related Work}
\vspace{-2mm}
We propose a novel approach to ZSL in the challenging setting where data arrives in a sequential manner. We cover here works in both ZSL and continual learning, as well as the few works that discuss both.

\vspace{-2mm}
\subsection{Zero-Shot Learning}
\vspace{-2mm}
The ZSL literature is both vast and diverse, with approaches that can be roughly divided into two categories: ($i$) non-generative and ($ii$)) generative approaches. Initial work~\cite{akata2013label,akata2015evaluation,norouzi2013zero,hwang2014unified,fu2015zero,xian2016latent} mainly focused on non-generative models. The objective of non-generative models is to learn a function from the seen classes that can measure the similarity between the visual and semantic spaces. \citep{akata2013label,norouzi2013zero,lampert2014attribute,xian2016latent} measure the linear compatibility between the visual and semantic spaces; modeling the complex relation, linear compatibility is not as prominent. Another set of works~\cite{saligram2016learningJoint,romera2015embarrassingly,kodirov2015unsupervisedDA}
focus on modeling relations by using bilinear compatibility relations, showing improved performance in similarity measure. These approaches show promising results for the ZSL setup where only unseen classes are evaluated on during test time, but when also simultaneously evaluated on seen classes ($i.e.$, GZSL), they perform poorly. This is primarily due to the inability to handle model bias towards the seen class samples. Recent works~\cite{skorokhodov2021class,liu2021isometric} have shown promising results for the GZSL setup using class normalization and isometric propagation networks. 
\citet{purushwalkam2019task} proposed a mechanism to use class attributes to gate modules that further process visual features for GZSL, which is reminiscent but different from the self-gating of attributes that we propose.

Of late, generative approaches have been among the most popular for GZSL. Because of the rapid progress in generative modeling ($e.g.$,  VAEs~\cite{kingma2014vae} and GANs~\cite{GAN,wgan}) generative approaches have been able to synthesize increasingly high-quality and realistic samples. For example, \citep{vermageneralized,verma2021towards,xian2018feature,mishra2017generative,felix2018multi,schonfeld2019generalized,chou2021adaptive,xian2019f,keshari2020generalized} have used conditional VAEs or GANs to generate samples for unseen classes conditioned on the class attribute. These synthesized samples can then be used for training alongside samples from the seen classes, transforming ZSL into traditional supervised learning.
Given the ability to generate as many samples as needed, these approaches can easily handle the model bias towards seen classes, leading to promising results for both ZSL and GZSL~\cite{verma2019meta,f-VAEGAN-D2}. 

\vspace{-2mm}
\subsection{Continual Learning}
\vspace{-2mm}
Catastrophic forgetting~\cite{mccloskey1989catastrophic,carpenter1987massively,lopez2017gradient,Kirkpatrick2017,jung2016less,rebuffi2017icarl,liang2018generative,rajasegaran2020itaml,kj2020meta} is a key problem for neural network learning from streams of data, with previous data no longer available. Continual learning methods seek to balance the goals of mitigating catastrophic forgetting of previous tasks, learning new tasks, and transferring knowledge from previous tasks forward to allow for quicker adaptation in the future. The continual learning literature is often broadly divided into three categories: ($i$) replay-based~\cite{rebuffi2017icarl,rajasegaran2020itaml,kj2020meta,van2018generative}, which rely on re-training the model with a small memory bank of samples from previous tasks; ($ii$) regularization-based~\cite{Kirkpatrick2017,yu2020semantic,lopez2017gradient,li2017learning}, which regularize the model parameters to minimize deviation of parameters important to previous tasks while learning novel tasks; and ($iii$) expansion-based models~\cite{rajasegaran2019random,masana2020ternary,xu2018reinforced,mallya2018piggyback,singh2020calibrating,mehta2021continual}, which increase model capacity dynamically with each new task, preserving parameters for previous tasks. 
While there are many evaluation protocols commonly used, class incremental learning~\cite{rajasegaran2020itaml,rebuffi2017icarl}, which does not assume a task identity (ID) during inference, is considered more realistic and challenging than the task incremental learning~\cite{singh2020calibrating,yoon2020scalable}, for which the task is known during inference.
{
	\renewcommand{\arraystretch}{1.4} 
\begin{table*}[ht]
	\centering
	\caption{Mean seen accuracy (mSA), mean unseen accuracy (mUA), and their harmonic mean (mH) for GZSL.}
	\addtolength{\tabcolsep}{-2pt}
	\resizebox{\textwidth}{!}{%
		\begin{tabular}{l|ccc|ccc|ccc|ccc|ccc}
			\hline
			& \multicolumn{3}{c|}{SUN} & \multicolumn{3}{c|}{CUB} & \multicolumn{3}{c|}{AWA1} & \multicolumn{3}{c|}{AWA2} & \multicolumn{1}{c}{Average}\\
			\cline{2-13}          
			& \multicolumn{1}{p{2.215em}}{mSA} & \multicolumn{1}{p{2.215em}}{mUA} & \multicolumn{1}{p{2.215em}|}{mH} & \multicolumn{1}{p{2.215em}}{mSA} & \multicolumn{1}{p{2.215em}}{mUA} & \multicolumn{1}{p{2.215em}|}{mH} & \multicolumn{1}{p{2.215em}}{mSA} & \multicolumn{1}{p{2.215em}}{mUA} & \multicolumn{1}{p{2.215em}|}{mH} & \multicolumn{1}{p{2.215em}}{mSA} & \multicolumn{1}{p{2.215em}}{mUA} & \multicolumn{1}{p{2.215em}|}{mH} & 
			\multicolumn{1}{p{6.em}}{Training Time} & \\
			\hline
			CVC-ZSL~\cite{li2019rethinking} & 36.3 &42.8& 39.3 &47.4& 47.6& 47.5& 62.7& 77.0& 69.1& 56.4& 81.4& 66.7& 3 Hours \\
			SGAL~\cite{yu2019zero} & 42.9& 31.2& 36.1& 47.1& 44.7& 45.9& 52.7& 75.7& 62.2& 55.1& 81.2& 65.6& 50 Min  \\
			SGMA~\cite{zhu2019semantic} & --& --& --& 36.7& 71.3& 48.5& -& -& -& 37.6& 87.1& 52.5& --  \\
			DASCN~\cite{ni2019dual} & 42.4 &38.5& 40.3& 45.9& 59.0& 51.6& 59.3& 68.0& 63.4& --& --& --& --  \\
			TF-VAEGAN~\cite{narayan2020latent} & 45.6& 40.7& 43.0& 52.8& 64.7& 58.1& --& --& --& 59.8& 75.1& 66.6& 1.75 Hours \\
			EPGN~\cite{yu2020episode} & --& -- &-- &52.0& 61.1& 56.2& 62.1& 83.4& 71.2& 52.6& 83.5& 64.6&--  \\
			DVBE~\cite{min2020domain} &45.0& 37.2& 40.7& 53.2& 60.2& 56.5& --& --& --& 63.6& 70.8& 67.0& --  \\
			LsrGAN~\cite{vyas2020leveraging} & 44.8& 37.7& 40.9& 48.1& 59.1& 53.0& --& --& --& 54.6& 74.6& 63.0& 1.25 Hours  \\
			F-VAEGAN-D2~\cite{xian2019f} & 45.1 &38.0& 41.3& 48.4& 60.1& 53.6& --& --& --& 57.6& 70.6& 63.5& --  \\
			ZSML~\cite{verma2019meta} & 45.1&21.7& 29.3& 60.0& 52.1& 55.7& 57.4& 71.1& 63.5& 58.9& 74.6& 65.8&3 Hours  \\
			NM-ZSL~\cite{skorokhodov2021class} & 44.7 &41.6& 43.1& 49.9& 50.7& 50.3& 63.1& 73.4& 67.8& 60.2& 77.1& 67.6& 1 Min  \\
			\hline
			MCZSL (Ours) & 40.3 &46.9& \textbf{43.4}& 57.2& 66.4& \textbf{61.4}& 78.9&64.6& \textbf{71.8}& 77.9& 67.1& \textbf{72.1}& \textbf{31 Second}  \\
			\hline
		\end{tabular}%
	}	
	\label{tab:gzsl}%
\end{table*}%
}

\vspace{-2mm}
\subsection{Zero-Shot Continual Learning}
\vspace{-2mm}
One of the desiderata of continual learning techniques is forward transfer of previous knowledge to future tasks, which may not be known ahead of time; similarly, GZSL approaches seek to adapt models to novel, unseen classes while still being able to classify the seen classes.
As such, there are clear connections between the two problem settings.
Some recent works~\cite{lopez2017gradient,wei2020lifelong,skorokhodov2021class,gautam2020generalized} have drawn increasing attention towards continual zero-shot learning (CZSL). For example, \citep{wei2020lifelong} considers a task incremental learning setting, where task ID for each sample is provided during train and test, an easier and perhaps less realistic setting than class incremental learning.
A-GEM~\cite{chaudhry2018efficient} proposed a regularization-based model to overcome catastrophic forgetting while maximizing the forward transfer. 
\citep{skorokhodov2021class} proposed a simple class normalization as an efficient solution to ZSL and extended it to CZSL; they proposed the setting discussed in Section~\ref{sec:setting-1}. 
Meanwhile, \citep{gautam2020generalized} proposed a replay-based approach for  CZSL, showing state-of-the-art results in a more realistic setting discussed in Section~\ref{sec:setting2}. Like \citep{gautam2020generalized,skorokhodov2021class}, we follow the class incremental learning scenario and evaluate our approach in both settings.

\section{Experiments}
\subsection{Training and Evaluation Protocols}
\subsubsection{Generalized Zero-Shot Learning (GZSL)}
The simplest case we consider is the generalized zero-shot learning (GZSL) setting~\cite{xian2018zero}.
In GZSL, classes are split into two groups: classes whose data are available during the model's training stage (``seen'' classes), and classes whose data only appear during inference (``unseen'' classes).
For both types, attribute vectors describing each class
are available to facilitate knowledge transfer.
During test time, samples may come from either classes seen during training or new unseen classes.
We report mean seen accuracy (mSA) and mean unseen accuracy (mUA), as well as the harmonic mean (mhM) of both as an overall metric; harmonic mean is considered preferable to simple arithmetic mean as an overall metric, as it prevents either term from dominating~\cite{xian2018zero}. 

Note that some GZSL approaches (notably, generative ones) assume that the list of unseen classes and their attribute vectors are available during the training stage, even if their data are not; this inherently restricts these models to these known unseen classes.
Conversely, our approach only requires the attributes of the seen classes.
Also, in contrast to the continual GZSL settings described below, all seen classes are assumed available simultaneously during training.

\subsubsection{Fixed Continual GZSL}
\label{sec:setting-1}
The setting proposed by \citet{skorokhodov2021class} divides all classes of the dataset into $K$ subsets, each corresponding to a task. 
For task $T_t$, the first $t$ of these subsets are considered the seen classes, while the rest are unseen; this results in the number of seen classes increasing with $t$ while the number of unseen classes decreases.
Over the span of $t=1,...K$, this simulates a scenario where we eventually ``collect'' labeled data for classes that were previously unseen.
Note that in contrast to the typical GZSL setting, only data from the $t^\mathrm{th}$ subset are available; previous training data are assumed inaccessible.
The goal is to learn from this newly ``collected'' data without experiencing catastrophic forgetting.
As in GZSL, we report mSA, mUA, and mH, but at the end of $K-1$ tasks: 
\begin{eqnarray}
mSA_F=&\frac{1}{K-1}\sum_{i=1}^{K-1}\mathrm{Acc}(\cD_{ts}^i(c_{\leq i}^s),\cA(c_{\leq i}^s))\\
mUA_F=&\frac{1}{K-1}\sum_{i=1}^{K-1}\mathrm{Acc}(\cD_{ts}^i(c_{i}^u),\cA(c_{i}^u))\\
mhM_F=&\frac{1}{K-1}\sum_{i=1}^{K-1}\cH(\cD_{ts}^i(c_{\leq i}^s),\cD_{ts}^i(c_{i}^u),\cA)
\end{eqnarray}
where $\mathrm{Acc}$ represents per class accuracy, $\cD_{ts}^i(c_{\leq i}^s)$ and $\cA(c_{\leq t}^s)$ are the seen class test data and attribute vectors respectively during the $i^{\mathrm{th}}$ task. Similarly $\cD_{ts}^i(c_{i}^u)$ and $\cA(c_{i}^u)$ represents the unseen class test data and attribute vectors during the $i^{\mathrm{th}}$ task. $\cH$ is the harmonic mean of the accuracies obtained on $\cD_{ts}^i(c_{\leq i}^s)$ and $\cD_{ts}^i(c_{i}^u)$. We calculate the metric up to task $K-1$, as there are no unseen classes for task $K$, resulting in standard supervised continual learning.

{
	\renewcommand{\arraystretch}{1.4} 
	\begin{table*}[ht]
		\centering
		\caption{Mean seen accuracy (mSA), mean unseen accuracy (mUA), and their harmonic mean (mH) for fixed continual GZSL.}
		\addtolength{\tabcolsep}{-3pt}
		\resizebox{\textwidth}{!}{%
			\begin{tabular}{l|ccc|ccc|ccc|ccc|ccc}
				\hline
				& \multicolumn{3}{c|}{CUB} & \multicolumn{3}{c|}{aPY} & \multicolumn{3}{c|}{AWA1} & \multicolumn{3}{c|}{AWA2} & \multicolumn{3}{c}{SUN} \\
				\cline{2-16}          & \multicolumn{1}{p{2.215em}}{mSA} & \multicolumn{1}{p{2.215em}}{mUA} & \multicolumn{1}{p{2.215em}|}{mH} & \multicolumn{1}{p{2.215em}}{mSA} & \multicolumn{1}{p{2.215em}}{mUA} & \multicolumn{1}{p{2.215em}|}{mH} & \multicolumn{1}{p{2.215em}}{mSA} & \multicolumn{1}{p{2.215em}}{mUA} & \multicolumn{1}{p{2.215em}|}{mH} & \multicolumn{1}{p{2.215em}}{mSA} & \multicolumn{1}{p{2.215em}}{mUA} & \multicolumn{1}{p{2.215em}|}{mH} & \multicolumn{1}{p{2.215em}}{mSA} & \multicolumn{1}{p{2.215em}}{mUA} & \multicolumn{1}{p{2.215em}}{mH} \\
				\hline
				Sequential (Lower Bound) & 11.44 & 2.84  & 4.25 & 36.53 & 15.78 & 17.66 & 40.91 & 12.08 & 18.11 & 43.37 & 12.02 & 18.12 & 11.82 & 3.06 & 4.78 \\
				Seq-CVAE~\cite{mishra2017generative} & 24.66 & 8.57  & 12.18 & 51.57 & 11.38 & 18.33 & 59.27 & 18.24 & 27.14 & 61.42 & 19.34 & 28.67 & 16.88 & 11.40 & 13.38 \\
				Seq-CADA~\cite{schonfeld2019generalized} & 40.82 & 14.37 & 21.14 & 45.25 & 10.59 & 16.42 & 51.57 & 18.02 & 27.59 & 52.30 & 20.30 & 30.38 & 25.94 & 16.22 & 20.10 \\
				AGEM-CZSL~\cite{chaudhry2018efficient} & --  & --  & 13.20 & --  & --  & --  & --  & --  & --  & --  & --  & --  & --  & --  & 10.50 \\
				CZSL-CV+res~\cite{gautam2020generalized} & 44.89 & 13.45 & 20.15 & 64.88 & 15.24 & 23.90 & 78.56 & 23.65 & 35.51 & 80.97 & 25.75 & 38.34 & 23.99 & 14.10 & 17.63 \\
				CZSL-CA+res~\cite{gautam2020generalized} & 43.96 & 32.77 & 36.06 & 57.69 & 20.83 & 28.84 & 62.64 & 38.41 & 45.38 & 62.80 & 39.23 & 46.22 & 27.11 & 21.72 & 22.92 \\
				NM-ZSL~\cite{skorokhodov2021class} & 55.45 & 43.25 & 47.04 & 45.26 & 21.35 & 27.18 & 70.90 & 37.46 & 48.75 & 76.33 & 39.79 & 51.51 & 50.01 & 19.77 & 28.04 \\
				\hline
				MCZSL (Ours) & 58.34 & 48.35  & \textbf{51.31} &50.74& 22.37 & \textbf{30.66} & 63.69 & 46.71 &\textbf{53.12}& 68.01 & 48.38 & \textbf{55.17}  & 55.40 & 22.91 & \textbf{32.23}  \\
				\hline
			\end{tabular}%
		}
		\label{tab:gen_res_S1}
	\end{table*}
}	

{
	\renewcommand{\arraystretch}{1.4} 
	\begin{table*}[ht]
		\centering
		\caption{Mean seen accuracy (mSA), mean unseen accuracy (mUA), and their harmonic mean (mH) for dynamic continual GZSL.}
		\addtolength{\tabcolsep}{-3pt}
		\resizebox{\textwidth}{!}{%
			\begin{tabular}{l|ccc|ccc|ccc|ccc|ccc}
				\hline
				& \multicolumn{3}{c|}{CUB} & \multicolumn{3}{c|}{aPY} & \multicolumn{3}{c|}{AWA1} & \multicolumn{3}{c|}{AWA2} & \multicolumn{3}{c}{SUN}\\
				\cline{2-16}          & \multicolumn{1}{p{2.215em}}{mSA} & \multicolumn{1}{p{2.215em}}{mUA} & \multicolumn{1}{p{2.215em}|}{mH} & \multicolumn{1}{p{2.215em}}{mSA} & \multicolumn{1}{p{2.215em}}{mUA} & \multicolumn{1}{p{2.215em}|}{mH} & \multicolumn{1}{p{2.215em}}{mSA} & \multicolumn{1}{p{2.215em}}{mUA} & \multicolumn{1}{p{2.215em}|}{mH} & \multicolumn{1}{p{2.215em}}{mSA} & \multicolumn{1}{p{2.215em}}{mUA} & \multicolumn{1}{p{2.215em}|}{mH} & \multicolumn{1}{p{2.215em}}{mSA} & \multicolumn{1}{p{2.215em}}{mUA} & \multicolumn{1}{p{2.215em}}{mH} \\
				\hline
				Sequential (Lower Bound) & 15.82 & 8.35  & 9.53 & 52.26 & 25.21 & 30.56 & 48.01 & 31.97 & 35.84 & 49.56 & 26.56 & 31.81 & 16.17 & 7.42 & 9.41 \\
				Seq-CVAE~\cite{mishra2017generative} & 38.95 & 20.89 & 26.74 & 65.87 & 17.90 & 25.84 & 70.24 & 28.36 & 39.32 & 73.71 & 26.22 & 36.30 & 29.06 & 21.33 & 24.33 \\
				Seq-CADA~\cite{schmidhuber1987evolutionary} & 55.55 & 26.96 & 35.62 & 61.17 & 21.13 & 26.37 & 78.12 & 35.93 & 47.06 & 79.89 & 36.64 & 47.99 & 42.21 & 23.47 & 29.60 \\
				CZSL-CV+res~\cite{gautam2020generalized} & 63.16 & 27.50 & 37.84 & 78.15 & 28.10 & 40.21 & 85.01 & 37.49 & 51.60 & 88.36 & 33.24 & 47.89 & 37.50 & 24.01 & 29.15\\
				CZSL-CA+res~\cite{gautam2020generalized} & 68.18 & 42.44 & 50.68 & 66.30 & 36.59 & \textbf{45.08} & 81.86 & 61.39 & 69.92 & 82.19 & 55.98 & 65.95 & 47.18 & 30.30 & 34.88 \\
				NM-ZSL~\cite{skorokhodov2021class} & 64.91 & 46.05 & 53.79 & 79.60 & 22.29 & 32.61 & 75.59 & 60.87 & 67.44 & 89.22 & 51.38 & 63.41 & 50.56 & 35.55 & 41.65 \\
				\hline
				MCZSL (Ours)  & 62.41 &67.63  & \textbf{64.71} & 69.91 & 34.36 & 43.03 & 81.81& 65.47 &\textbf{72.22}& 86.09 &64.62& \textbf{72.96} & 52.73 & 41.78 &\textbf{46.45}\\
				\hline
			\end{tabular}%
		}	
		\label{tab:gen_res_S2}%
	\end{table*}%
}	

\begin{figure*}
    \centering
    \includegraphics[scale=0.43]{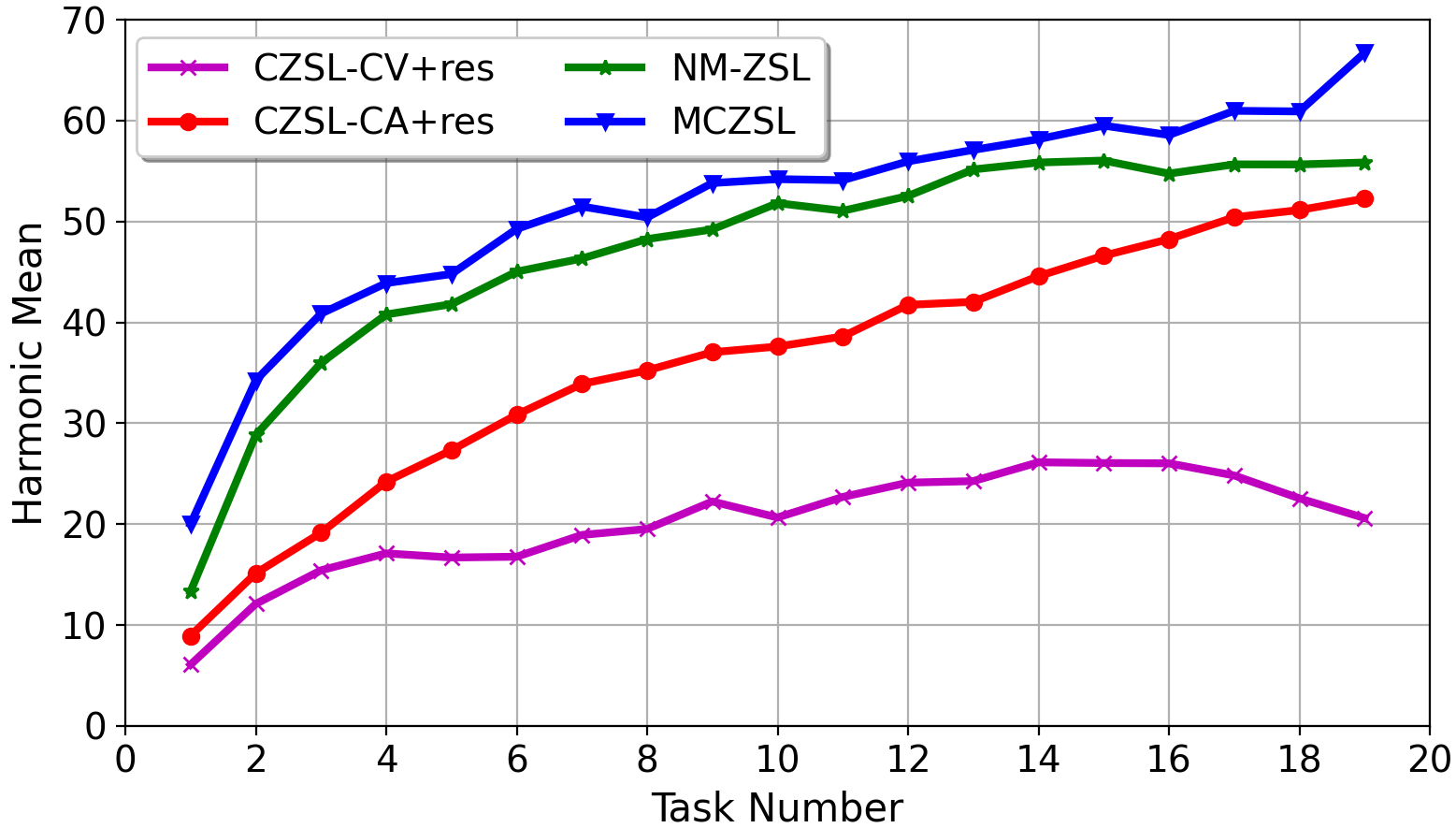}
    \includegraphics[scale=0.43]{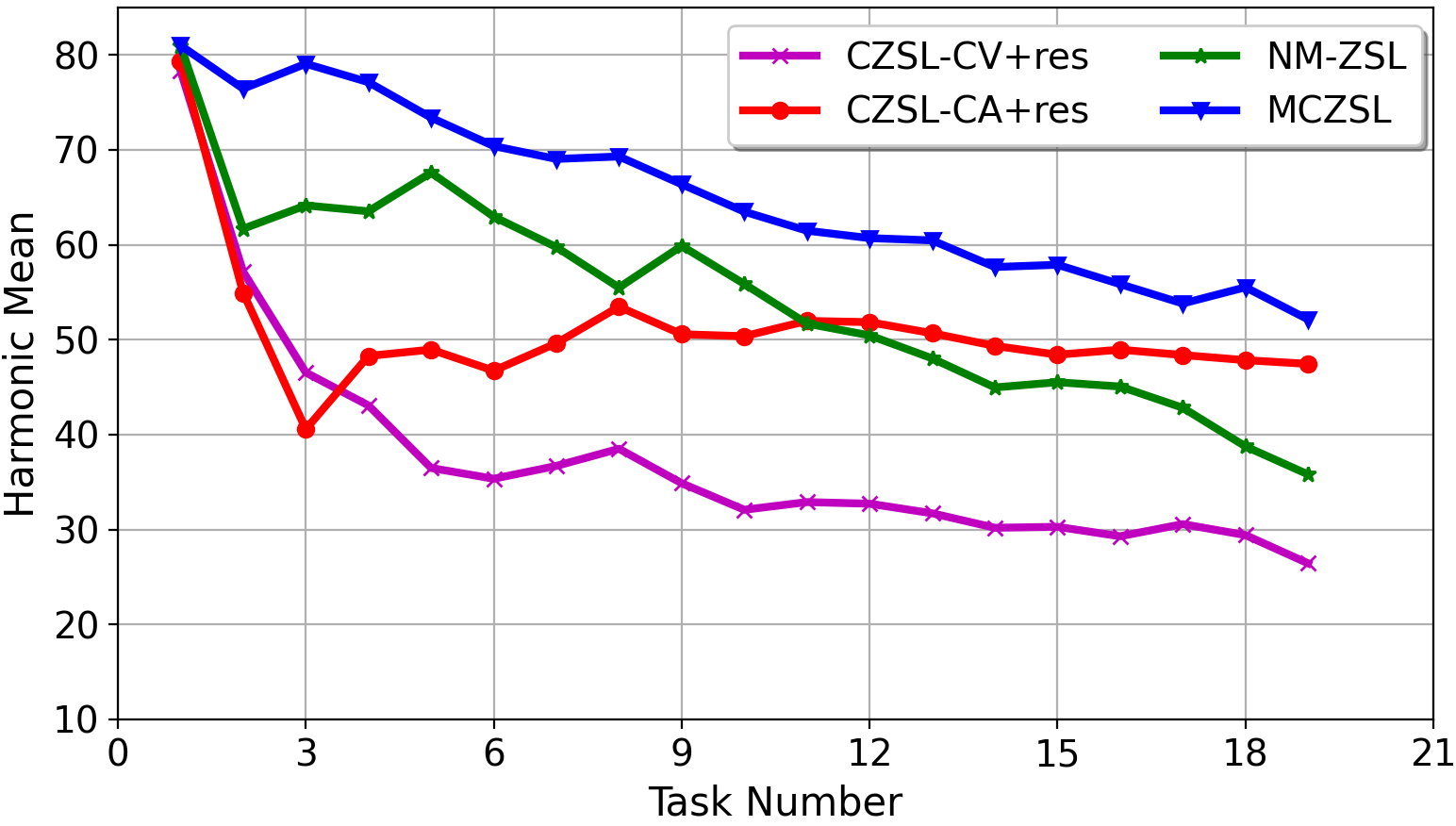}
    \caption{Harmonic mean per task for continual GZSL on the CUB dataset, Left: Fixed, Right: Dynamic}
    \label{fig:hmean_pertask}
\end{figure*}

\begin{figure*}[ht]
    \centering
    \includegraphics[scale=0.43]{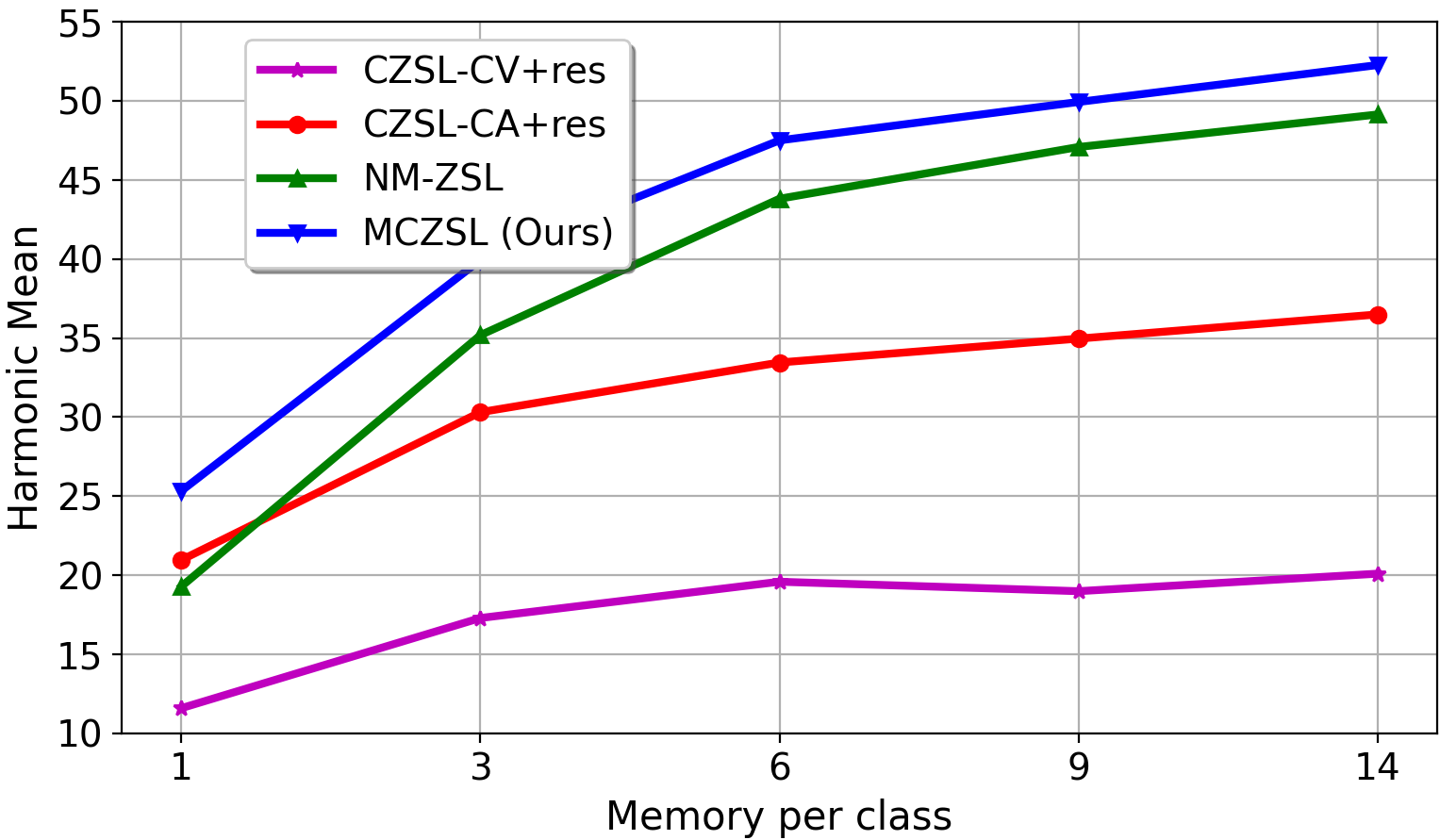} \quad
    \includegraphics[scale=0.43]{setting_1_Abalation_Mem.png}
    \caption{Memory growth vs model performance for continual GZSL on the CUB dataset, Left: Fixed, Right: Dynamic}
    \label{fig:memoryvs_hmean}
\end{figure*}

\subsubsection{Dynamic Continual GZSL}
\label{sec:setting2}
While it's not unreasonable that previously unseen class may become seen in the future, the above fixed continual GZSL evaluation protocol assumes that all unseen classes and attributes are set from the beginning, which may be unrealistic.
An alternative framing of continual GZSL is one in which each class consists of its own disjoint set of seen and unseen classes, as proposed by \citet{gautam2020generalized}.
Such a formulation does not require all attributes to be known \textit{a priori}, allowing the model to continue accommodating an unbounded number of classes.
As such, in contrast to the fixed continual GZSL, the number of seen and unseen classes both increase with $t$.
As with the other settings, we report mSA, mUA and mH:
\begin{eqnarray}
    mSA_D=&\frac{1}{K}\sum_{i=1}^{K}\mathrm{Acc}(\cD_{ts}^i(c_{\leq i}^s),\cA(c_{\leq i}^s))\\
    mUA_D=&\frac{1}{K}\sum_{i=1}^{K}\mathrm{Acc}(\cD_{ts}^i(c_{\leq i}^u),\cA(c_{\leq i}^u))\\
    mhM_D=&\frac{1}{K}\sum_{i=1}^{K}\cH(\cD_{ts}^i(c_{\leq i}^s),\cD_{ts}^i(c_{\leq i}^u),\cA)
\end{eqnarray}
where $\mathrm{Acc}$ represents per class accuracy, $\cD_{ts}^i(c_{\leq i}^s)$ and $\cA(c_{\leq t}^s)$ are the seen class test data and attribute vectors during $i^{\mathrm{th}}$ task. Similarly $\cD_{ts}^i(c_{\leq i}^u)$ and $\cA(c_{\leq i}^u)$ represents the unseen class test data and attribute vector during the $i^{\mathrm{th}}$ task. Detailed splits of the seen and unseen class samples for each task are given in the supplementary material.

\subsection{Datasets and Baselines}
\paragraph{Datasets} We conduct experiments on five widely used datasets for zero-shot learning. CUB-200~\cite{CUB} is a fine-grain dataset containing 200 classes of birds, and AWA1~\cite{AWA1} and AWA2~\cite{xian2018zero} are datasets containing 50 classes of animal, each represented by a $85$-dimensional attribute. aPY~\cite{aPY} is a diverse dataset containing 32 classes, each associated with a 64-dimensional attribute. SUN~\cite{patterson2012sun} contains $717$ classes, each with only 20 samples; fewer samples and a high number of classes make SUN especially challenging. In the SUN dataset each class is represented by a 102-dimensional attribute vector. More dataset descriptions, hyperparameters, and implementation details can be found in the supplementary material.

\paragraph{Baselines}
We compare the proposed approach against a variety of baselines.
CZSL-CV+res and CZSL-CA+res~\cite{gautam2020generalized} are continual ZSL models incorporating a conditional variational auto-encoders (VAE) and CADA~\cite{schonfeld2019generalized} with a memory reservoir, respectively. AGEM-CZSL~\cite{chaudhry2018efficient} is the average gradient episodic memory-based continual ZSL method. 
NM-ZSL~\cite{skorokhodov2021class} is an embedding-based approach that originally proposed what we refer to as fixed GZSL.
Seq-CVAE and Seq-CADA are the sequential versions of VAE~\cite{mishra2017generative} and CADA~\cite{schonfeld2019generalized} for CZSL settings, which can be considered as a lower bound of a generative approach. 

\subsection{Results}
\paragraph{Generalized Zero-Shot Learning}
We conduct experiments on CUB-200, SUN, AWA1, and AWA2 datasets in the GZSL setting, reporting means of the seen and unseen classes, as well as their harmonic means, in Table~\ref{tab:gzsl}.
We observe significant improvements over previous methods across all datasets, with absolute gains of $0.3\%$, $2.3\%$, $0.6\%$, and $4.5\%$ for SUN, CUB, AWA1, and AWA2 dataset. 
Particularly noteworthy is the average training\footnote{Time to train on the whole dataset with a Nvidia GTX 1080Ti} time of our approach relative to the baseline approaches, many of which require learning complex generative models ($e.g.$, VAE or GAN) to synthesize the realistic samples.
Compared to other generative approaches, MCZSL can be $100$-$300\times$ faster. 
 
\paragraph{Fixed Continual GZSL}
We perform fixed continual GZSL experiments on the CUB, aPY, AWA1, AWA2 and SUN datasets, showing the results in Table~\ref{tab:gen_res_S1}. 
For fair comparison, we use the same memory buffer size as \citet{gautam2020generalized}. 
On CUB, aPY, AWA1, AWA2 and SUN, the proposed model shows $4.27\%$, $1.82\%$, $4.37\%$, $3.66\%$ and $4.19\%$ absolute increase over the best baseline.
Figure~\ref{fig:hmean_pertask} (left) shows harmonic mean of the model versus the task number for the CUB dataset.
With each task, more of the unseen classes become seen, so performance tends to increase over time.
We observe that our proposed MCZSL consistently outperforms recent baselines.   
Notably, these significant improvements are achieved without using any costly generative models. 
Detailed descriptions of the experimental setup, task splits in each class, and memory reservoir size are given in the supplementary material. 

\begin{figure}
    \centering
    \includegraphics[scale=0.32]{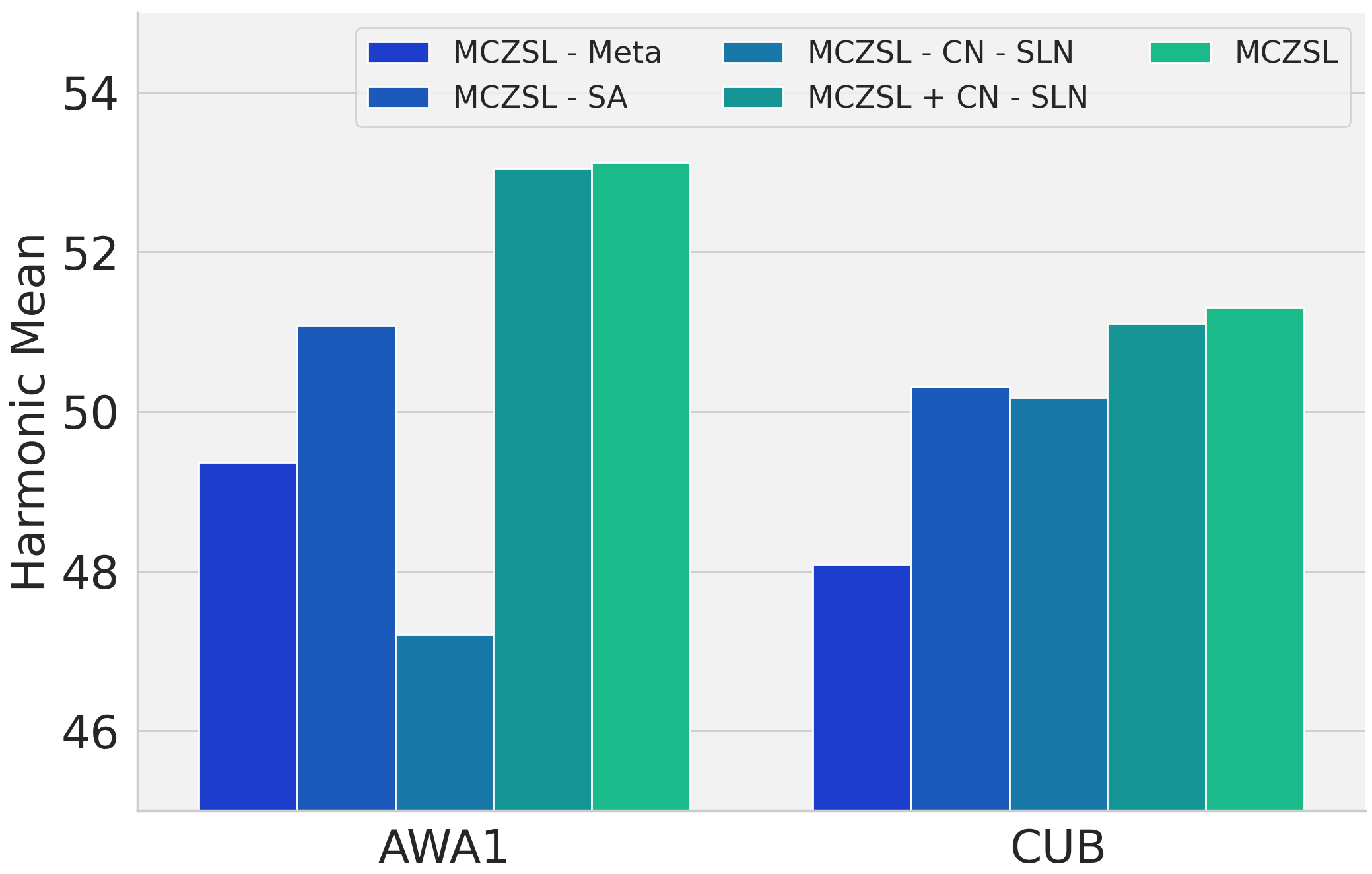}
    \caption{Effect of the different proposed components on the CUB and AWA1 datasets for fixed continual GZSL. \emph{MCZSL}: our model with all components, \emph{MCZSL-meta}: without meta-learning module, \emph{MCZSL-SA}: without self-gating, \emph{MCZSL-CN-SN}: without CN or SCN, and \emph{MCZSL+CN-SCL}: with CN but without SCN.}
    \label{fig:self_attention}
\end{figure}

\paragraph{Dynamic Continual GZSL}
As with fixed continual GZSL, we also conduct experiments in the dynamic continual GZSL on CUB, aPY, AWA1, AWA2 and SUN as shown in Table~\ref{tab:gen_res_S2}.
Once again, we observe that our proposed MCZSL performs well, seeing absolute gains of $10.92\%$, $2.30\%$, $7.01\%$, and $4.80\%$, on the CUB, AWA1, AWA2, and SUN datasets, while being relatively competitive on aPY. In Figure~\ref{fig:hmean_pertask} (right), we show harmonic mean per task for the CUB dataset.
In contrast with fixed continual GZSL, each task brings more seen and unseen classes, making the problem harder due to more classes to distinguish and increasing opportunity for forgetting; thus, accuracy tends to drop with more tasks.
Regardless, we again observe that the proposed model consistently outperforms recent baselines over the task numbers.

\subsection{Ablation Studies}
We conduct extensive ablation studies on the different components of the proposed model, observing that each of the proposed components play a critical role. We show the effects of different components on the AWA1 and CUB datasets in the fixed continual GZSL setting, with more ablation studies for dynamic continual GZSL in the supplementary material.

\subsubsection{Reservoir size vs Performance}
To overcome catastrophic forgetting, the model uses a constant-size reservoir~\cite{lopez2017gradient} to store previous task samples; with more tasks, the number of samples per class decreases. The reservoir size plays a key role for model performance. In Figure~\ref{fig:memoryvs_hmean}, we evaluate the model's performance for both fixed and dynamic continual GZSL. We observe that for different reservoir sizes $\{1,3,6,9,14\} \times \# classes$, the proposed model shows consistently better results compared to recent models. For fixed and dynamic continual GZSL, $\# classes$ is $S+U$ and $S$ respectively. 

\subsubsection{Effect of Self-Gating on the Attribute}
We apply self-gating to the learned embedding space of the attribute vector to get a robust and global representation. 
Ablations in Figure~\ref{fig:self_attention} show the effect of self-gating. 
We observe that our proposed MCZSL achieves harmonic means of $51.31$ and $53.12$ for the CUB and AWA1 datasets, respectively, which drop to $50.29$ and $51.08$ without self-gating.

\subsubsection{Effect of Scaled Class Normalization}
We observe that class normalization also plays a significant role to improve the model's performance. Removing the class normalization drops the harmonic mean from $53.12$ to $47.21$ and $51.31$ to $50.17$ for the AWA1 and CUB datasets respectively. Scaling the $\mu$ and $\sigma$ using a learnable parameter $\alpha$ and $\beta$ (as in (\ref{eq:scaleCN})) slightly improves the model's performance. Refer to the Figure~\ref{fig:self_attention} for more details.

\subsubsection{Effect of Meta-training}
Because of its learning ability using the few samples and quick adaption, meta-learning~\cite{nichol2018first,finn2017model} based training plays a crucial role to improve the model performance. The meta-learning framework are described in the Section~\ref{sec:meta}. We evaluate the model performance with and without meta-learning based training for fixed continual learning. The result are shown in the Figure~\ref{fig:self_attention}. We empirically observe that if we withdraw the meta learning (MCZSL - meta) based training, the MCZSL performance drops significantly. On the AWA1 and CUB datasets in the fixed continual GZSL setting, the harmonic mean of MCZSL drops from $53.12$ to $49.36$ and $51.31$ to $48.08$ respectively.

\section{Conclusions}
We have proposed meta-continual zero-shot learning (MCZSL), a zero-shot learning method capable of operating in both generalized and continual learning settings.
Through a novel self-gating on attributes and a scaled layer normalization, we obtain state-of-the art results in GZSL settings, despite not using expensive generative models; this allows for considerably faster speed during training, as well as flexibility to generalize to unlimited unseen classes that were unknown during training time.
Given clear connections between zero-shot and continual learning, we also extend our approach to settings where data arrives sequentially.
We adopt a data reservoir approach to mitigate catastrophic forgetting, and couple it with a meta-learning few-shot based approach to naturally enable the model to efficiently learn from the few samples that can be saved in a buffer.
Our results on the CUB, aPY, AWA1, AWA2 and SUN datasets in GZSL and two different protocols of continual GZSL demonstrate that our approach outperforms a wide array of strong, recent baselines.
Ablation studies demonstrate that each component of our proposed approach is critical for its success.

\bibliography{egbib}

\begin{thebibliography}{73}
\providecommand{\natexlab}[1]{#1}
\providecommand{\url}[1]{\texttt{#1}}
\expandafter\ifx\csname urlstyle\endcsname\relax
  \providecommand{\doi}[1]{doi: #1}\else
  \providecommand{\doi}{doi: \begingroup \urlstyle{rm}\Url}\fi

\bibitem[Akata et~al.(2013)Akata, Perronnin, Harchaoui, and
  Schmid]{akata2013label}
Akata, Z., Perronnin, F., Harchaoui, Z., and Schmid, C.
\newblock Label-embedding for attribute-based classification.
\newblock In \emph{Proceedings of the IEEE/CVF Conference on Computer Vision
  and Pattern Recognition}, pp.\  819--826, 2013.

\bibitem[Akata et~al.(2015)Akata, Reed, Walter, Lee, and
  Schiele]{akata2015evaluation}
Akata, Z., Reed, S., Walter, D., Lee, H., and Schiele, B.
\newblock Evaluation of output embeddings for fine-grained image
  classification.
\newblock In \emph{Proceedings of the IEEE/CVF Conference on Computer Vision
  and Pattern Recognition}, pp.\  2927--2936, 2015.

\bibitem[Arjovsky et~al.(2017)Arjovsky, Chintala, and Bottou]{wgan}
Arjovsky, M., Chintala, S., and Bottou, L.
\newblock Wasserstein gan.
\newblock \emph{arXiv preprint arXiv:1701.07875}, 2017.

\bibitem[Ba et~al.(2016)Ba, Kiros, and Hinton]{ba2016layer}
Ba, J.~L., Kiros, J.~R., and Hinton, G.~E.
\newblock Layer normalization.
\newblock \emph{arXiv preprint arXiv:1607.06450}, 2016.

\bibitem[Carpenter \& Grossberg(1987)Carpenter and
  Grossberg]{carpenter1987massively}
Carpenter, G.~A. and Grossberg, S.
\newblock A massively parallel architecture for a self-organizing neural
  pattern recognition machine.
\newblock \emph{Computer vision, graphics, and image processing}, pp.\
  54--115, 1987.

\bibitem[Chaudhry et~al.(2019)Chaudhry, Ranzato, Rohrbach, and
  Elhoseiny]{chaudhry2018efficient}
Chaudhry, A., Ranzato, M., Rohrbach, M., and Elhoseiny, M.
\newblock Efficient lifelong learning with a-gem.
\newblock In \emph{International Conference on Learning Representations}, 2019.

\bibitem[Chou et~al.(2021)Chou, Lin, and Liu]{chou2021adaptive}
Chou, Y.-Y., Lin, H.-T., and Liu, T.-L.
\newblock Adaptive and generative zero-shot learning.
\newblock In \emph{International Conference on Learning Representations}, 2021.
\newblock URL \url{https://openreview.net/forum?id=ahAUv8TI2Mz}.

\bibitem[Farhadi et~al.(2009)Farhadi, Endres, Hoiem, and Forsyth]{aPY}
Farhadi, A., Endres, I., Hoiem, D., and Forsyth, D.
\newblock Describing objects by their attributes.
\newblock In \emph{2009 IEEE Conference on Computer Vision and Pattern
  Recognition}, pp.\  1778--1785. IEEE, 2009.

\bibitem[Felix et~al.(2018)Felix, Reid, Carneiro, et~al.]{felix2018multi}
Felix, R., Reid, I., Carneiro, G., et~al.
\newblock Multi-modal cycle-consistent generalized zero-shot learning.
\newblock In \emph{Proceedings of the European Conference on Computer Vision
  (ECCV)}, pp.\  21--37, 2018.

\bibitem[Finn et~al.(2017)Finn, Abbeel, and Levine]{finn2017model}
Finn, C., Abbeel, P., and Levine, S.
\newblock Model-agnostic meta-learning for fast adaptation of deep networks.
\newblock In \emph{International Conference on Machine Learning}, pp.\
  1126--1135. PMLR, 2017.

\bibitem[Fu et~al.(2015)Fu, Xiang, Kodirov, and Gong]{fu2015zero}
Fu, Z., Xiang, T., Kodirov, E., and Gong, S.
\newblock Zero-shot object recognition by semantic manifold distance.
\newblock In \emph{Proceedings of the IEEE/CVF Conference on Computer Vision
  and Pattern Recognition}, pp.\  2635--2644, 2015.

\bibitem[Gautam et~al.(2020)Gautam, Parameswaran, Mishra, and
  Sundaram]{gautam2020generalized}
Gautam, C., Parameswaran, S., Mishra, A., and Sundaram, S.
\newblock Generalized continual zero-shot learning.
\newblock \emph{arXiv preprint arXiv:2011.08508}, 2020.

\bibitem[Goodfellow et~al.(2014)Goodfellow, Pouget-Abadie, Mirza, Xu,
  Warde-Farley, Ozair, Courville, and Bengio]{GAN}
Goodfellow, I., Pouget-Abadie, J., Mirza, M., Xu, B., Warde-Farley, D., Ozair,
  S., Courville, A., and Bengio, Y.
\newblock Generative adversarial nets.
\newblock In \emph{Advances in Neural Information Processing Systems}, pp.\
  2672--2680, 2014.

\bibitem[He et~al.(2016)He, Zhang, Ren, and Sun]{he2016deep}
He, K., Zhang, X., Ren, S., and Sun, J.
\newblock Deep residual learning for image recognition.
\newblock In \emph{Proceedings of the IEEE conference on computer vision and
  pattern recognition}, pp.\  770--778, 2016.

\bibitem[Hwang \& Sigal(2014)Hwang and Sigal]{hwang2014unified}
Hwang, S.~J. and Sigal, L.
\newblock A unified semantic embedding: Relating taxonomies and attributes.
\newblock \emph{Advances in Neural Information Processing Systems}, 2014.

\bibitem[Ioffe \& Szegedy(2015)Ioffe and Szegedy]{ioffe2015batch}
Ioffe, S. and Szegedy, C.
\newblock Batch normalization: Accelerating deep network training by reducing
  internal covariate shift.
\newblock In \emph{International conference on machine learning}, pp.\
  448--456. PMLR, 2015.

\bibitem[Jung et~al.(2016)Jung, Ju, Jung, and Kim]{jung2016less}
Jung, H., Ju, J., Jung, M., and Kim, J.
\newblock Less-forgetting learning in deep neural networks.
\newblock \emph{arXiv preprint arXiv:1607.00122}, 2016.

\bibitem[Keshari et~al.(2020)Keshari, Singh, and Vatsa]{keshari2020generalized}
Keshari, R., Singh, R., and Vatsa, M.
\newblock Generalized zero-shot learning via over-complete distribution.
\newblock In \emph{Proceedings of the IEEE/CVF Conference on Computer Vision
  and Pattern Recognition}, pp.\  13300--13308, 2020.

\bibitem[Kingma \& Ba(2014)Kingma and Ba]{kingma2014adam}
Kingma, D. and Ba, J.
\newblock Adam: A method for stochastic optimization.
\newblock \emph{arXiv preprint arXiv:1412.6980}, 2014.

\bibitem[Kingma \& Welling(2014)Kingma and Welling]{kingma2014vae}
Kingma, D.~P. and Welling, M.
\newblock Auto-encoding variational bayes.
\newblock In \emph{International Conference on Learning Representations}, 2014.

\bibitem[Kirkpatrick et~al.(2017)Kirkpatrick, Pascanu, Rabinowitz, Veness,
  Desjardins, Rusu, Milan, Quan, Ramalho, Grabska-Barwinska, Hassabis, Clopath,
  Kumaran, and Hadsell]{Kirkpatrick2017}
Kirkpatrick, J., Pascanu, R., Rabinowitz, N., Veness, J., Desjardins, G., Rusu,
  A.~A., Milan, K., Quan, J., Ramalho, T., Grabska-Barwinska, A., Hassabis, D.,
  Clopath, C., Kumaran, D., and Hadsell, R.
\newblock {Overcoming Catastrophic Forgetting in Neural Networks}.
\newblock \emph{Proceedings of the National Academy of Sciences}, 2017.

\bibitem[KJ \& Nallure~Balasubramanian(2020)KJ and
  Nallure~Balasubramanian]{kj2020meta}
KJ, J. and Nallure~Balasubramanian, V.
\newblock {Meta-Consolidation for Continual Learning}.
\newblock \emph{Advances in Neural Information Processing Systems}, 33, 2020.

\bibitem[Kodirov et~al.(2015)Kodirov, Xiang, Fu, and
  Gong]{kodirov2015unsupervisedDA}
Kodirov, E., Xiang, T., Fu, Z., and Gong, S.
\newblock Unsupervised domain adaptation for zero-shot learning.
\newblock In \emph{The IEEE International Conference on Computer Vision}, pp.\
  2452--2460, 2015.

\bibitem[Krizhevsky et~al.(2012)Krizhevsky, Sutskever, and
  Hinton]{krizhevsky2012imagenet}
Krizhevsky, A., Sutskever, I., and Hinton, G.~E.
\newblock Imagenet classification with deep convolutional neural networks.
\newblock In \emph{Advances in Neural Information Processing Systems}, pp.\
  1097--1105, 2012.

\bibitem[Lampert et~al.(2009)Lampert, Nickisch, and Harmeling]{AWA1}
Lampert, C.~H., Nickisch, H., and Harmeling, S.
\newblock Learning to detect unseen object classes by between-class attribute
  transfer.
\newblock In \emph{2009 IEEE Conference on Computer Vision and Pattern
  Recognition}, pp.\  951--958. IEEE, 2009.

\bibitem[Lampert et~al.(2014)Lampert, Nickisch, and
  Harmeling]{lampert2014attribute}
Lampert, C.~H., Nickisch, H., and Harmeling, S.
\newblock Attribute-based classification for zero-shot visual object
  categorization.
\newblock \emph{IEEE Transactions on Pattern Analysis and Machine
  Intelligence}, 36\penalty0 (3):\penalty0 453--465, 2014.

\bibitem[Li et~al.(2019)Li, Min, and Fu]{li2019rethinking}
Li, K., Min, M.~R., and Fu, Y.
\newblock Rethinking zero-shot learning: A conditional visual classification
  perspective.
\newblock In \emph{Proceedings of the IEEE International Conference on Computer
  Vision}, pp.\  3583--3592, 2019.

\bibitem[Li \& Hoiem(2017)Li and Hoiem]{li2017learning}
Li, Z. and Hoiem, D.
\newblock {Learning without Forgetting}.
\newblock \emph{IEEE Transactions on Pattern Analysis and Machine
  Intelligence}, 2017.

\bibitem[Liang et~al.(2018)Liang, Li, Wang, and Carin]{liang2018generative}
Liang, K.~J., Li, C., Wang, G., and Carin, L.
\newblock {Generative Adversarial Network Training is a Continual Learning
  Problem}.
\newblock \emph{arXiv preprint arXiv:1811.11083}, 2018.

\bibitem[Liu et~al.(2021)Liu, Zhou, Long, Jiang, Dong, and
  Zhang]{liu2021isometric}
Liu, L., Zhou, T., Long, G., Jiang, J., Dong, X., and Zhang, C.
\newblock Isometric propagation network for generalized zero-shot learning.
\newblock In \emph{International Conference on Learning Representations}, 2021.
\newblock URL \url{https://openreview.net/forum?id=-mWcQVLPSPy}.

\bibitem[Lopez-Paz et~al.(2017)Lopez-Paz, Ranzato, and
  Marc'Aurelio]{lopez2017gradient}
Lopez-Paz, D., Ranzato, and Marc'Aurelio.
\newblock Gradient episodic memory for continual learning.
\newblock In \emph{Advances in Neural Information Processing Systems}, pp.\
  6467--6476, 2017.

\bibitem[Mallya et~al.(2018)Mallya, Davis, and Lazebnik]{mallya2018piggyback}
Mallya, A., Davis, D., and Lazebnik, S.
\newblock Piggyback: Adapting a single network to multiple tasks by learning to
  mask weights.
\newblock In \emph{Proceedings of the European Conference on Computer Vision
  (ECCV)}, pp.\  67--82, 2018.

\bibitem[Masana et~al.(2020)Masana, Tuytelaars, and van~de
  Weijer]{masana2020ternary}
Masana, M., Tuytelaars, T., and van~de Weijer, J.
\newblock Ternary feature masks: continual learning without any forgetting.
\newblock \emph{arXiv preprint arXiv:2001.08714}, 2020.

\bibitem[McCloskey \& Cohen(1989)McCloskey and
  Cohen]{mccloskey1989catastrophic}
McCloskey, M. and Cohen, N.~J.
\newblock Catastrophic interference in connectionist networks: The sequential
  learning problem.
\newblock In \emph{Psychology of learning and motivation}, volume~24, pp.\
  109--165. Elsevier, 1989.

\bibitem[Mehta et~al.(2021)Mehta, Liang, Verma, and Carin]{mehta2021continual}
Mehta, N., Liang, K.~J., Verma, V.~K., and Carin, L.
\newblock {Continual Learning using a Bayesian Nonparametric Dictionary of
  Weight Factors}.
\newblock \emph{Artificial Intelligence and Statistics}, 2021.

\bibitem[Min et~al.(2020)Min, Yao, Xie, Wang, Zha, and Zhang]{min2020domain}
Min, S., Yao, H., Xie, H., Wang, C., Zha, Z.-J., and Zhang, Y.
\newblock Domain-aware visual bias eliminating for generalized zero-shot
  learning.
\newblock In \emph{Proceedings of the IEEE/CVF Conference on Computer Vision
  and Pattern Recognition}, pp.\  12664--12673, 2020.

\bibitem[Mishra et~al.(2017)Mishra, Reddy, Mittal, and
  Murthy]{mishra2017generative}
Mishra, A., Reddy, M., Mittal, A., and Murthy, H.~A.
\newblock A generative model for zero shot learning using conditional
  variational autoencoders.
\newblock \emph{Proceedings of the IEEE/CVF Conference on Computer Vision and
  Pattern Recognition Workshop}, 2017.

\bibitem[Narayan et~al.(2020)Narayan, Gupta, Khan, Snoek, and
  Shao]{narayan2020latent}
Narayan, S., Gupta, A., Khan, F.~S., Snoek, C.~G., and Shao, L.
\newblock Latent embedding feedback and discriminative features for zero-shot
  classification.
\newblock \emph{arXiv preprint arXiv:2003.07833}, 2020.

\bibitem[Ni et~al.(2019)Ni, Zhang, and Xie]{ni2019dual}
Ni, J., Zhang, S., and Xie, H.
\newblock Dual adversarial semantics-consistent network for generalized
  zero-shot learning.
\newblock In \emph{Advances in Neural Information Processing Systems}, pp.\
  6146--6157, 2019.

\bibitem[Nichol et~al.(2018)Nichol, Achiam, and Schulman]{nichol2018first}
Nichol, A., Achiam, J., and Schulman, J.
\newblock On first-order meta-learning algorithms.
\newblock \emph{arXiv preprint arXiv:1803.02999}, 2018.

\bibitem[Norouzi et~al.(2013)Norouzi, Mikolov, Bengio, Singer, Shlens, Frome,
  Corrado, and Dean]{norouzi2013zero}
Norouzi, M., Mikolov, T., Bengio, S., Singer, Y., Shlens, J., Frome, A.,
  Corrado, G.~S., and Dean, J.
\newblock Zero-shot learning by convex combination of semantic embeddings.
\newblock \emph{arXiv preprint arXiv:1312.5650}, 2013.

\bibitem[Patterson \& Hays(2012)Patterson and Hays]{patterson2012sun}
Patterson, G. and Hays, J.
\newblock Sun attribute database: Discovering, annotating, and recognizing
  scene attributes.
\newblock In \emph{2012 IEEE Conference on Computer Vision and Pattern
  Recognition}, pp.\  2751--2758. IEEE, 2012.

\bibitem[Purushwalkam et~al.(2019)Purushwalkam, Nickel, Gupta, and
  Ranzato]{purushwalkam2019task}
Purushwalkam, S., Nickel, M., Gupta, A., and Ranzato, M.
\newblock Task-driven modular networks for zero-shot compositional learning.
\newblock In \emph{Proceedings of the IEEE/CVF International Conference on
  Computer Vision}, pp.\  3593--3602, 2019.

\bibitem[Rajasegaran et~al.(2019)Rajasegaran, Hayat, Khan, Khan, and
  Shao]{rajasegaran2019random}
Rajasegaran, J., Hayat, M., Khan, S.~H., Khan, F.~S., and Shao, L.
\newblock {Random Path Selection for Continual Learning}.
\newblock \emph{Advances in Neural Information Processing Systems}, 2019.

\bibitem[Rajasegaran et~al.(2020)Rajasegaran, Khan, Hayat, Khan, and
  Shah]{rajasegaran2020itaml}
Rajasegaran, J., Khan, S., Hayat, M., Khan, F.~S., and Shah, M.
\newblock itaml: An incremental task-agnostic meta-learning approach.
\newblock \emph{arXiv preprint arXiv:2003.11652}, 2020.

\bibitem[Rebuffi et~al.(2017)Rebuffi, Kolesnikov, Sperl, and
  Lampert]{rebuffi2017icarl}
Rebuffi, S.-A., Kolesnikov, A., Sperl, G., and Lampert, C.~H.
\newblock {iCaRL: Incremental Classifier and Representation Learning}.
\newblock \emph{Proceedings of the IEEE/CVF Conference on Computer Vision and
  Pattern Recognition}, 2017.

\bibitem[Romera \& Torr(2015)Romera and Torr]{romera2015embarrassingly}
Romera, Paredes, B. and Torr, P.~H.
\newblock An embarrassingly simple approach to zero-shot learning.
\newblock In \emph{International Conference on Machine Learning}, pp.\
  2152--2161, 2015.

\bibitem[Russakovsky et~al.(2015)Russakovsky, Deng, Su, Krause, Satheesh, Ma,
  Huang, Karpathy, Khosla, Bernstein, et~al.]{imagenet2015}
Russakovsky, O., Deng, J., Su, H., Krause, J., Satheesh, S., Ma, S., Huang, Z.,
  Karpathy, A., Khosla, A., Bernstein, M., et~al.
\newblock Imagenet large scale visual recognition challenge.
\newblock \emph{IJCV}, pp.\  211--252, 2015.

\bibitem[Schmidhuber(1987)]{schmidhuber1987evolutionary}
Schmidhuber, J.
\newblock \emph{Evolutionary principles in self-referential learning, or on
  learning how to learn: the meta-meta-... hook}.
\newblock PhD thesis, Technische Universit{\"a}t M{\"u}nchen, 1987.

\bibitem[Schonfeld et~al.(2019)Schonfeld, Ebrahimi, Sinha, Darrell, and
  Akata]{schonfeld2019generalized}
Schonfeld, E., Ebrahimi, S., Sinha, S., Darrell, T., and Akata, Z.
\newblock Generalized zero-and few-shot learning via aligned variational
  autoencoders.
\newblock In \emph{Proceedings of the IEEE Conference on Computer Vision and
  Pattern Recognition}, pp.\  8247--8255, 2019.

\bibitem[Singh et~al.(2020)Singh, Verma, Mazumder, Carin, and
  Rai]{singh2020calibrating}
Singh, P., Verma, V.~K., Mazumder, P., Carin, L., and Rai, P.
\newblock Calibrating cnns for lifelong learning.
\newblock \emph{Advances in Neural Information Processing Systems}, 33, 2020.

\bibitem[Skorokhodov et~al.(2021)Skorokhodov, Mohamed, and
  Elhoseiny]{skorokhodov2021class}
Skorokhodov, I., Mohamed, and Elhoseiny.
\newblock Class normalization for zero-shot learning.
\newblock In \emph{International Conference on Learning Representations}, 2021.
\newblock URL \url{https://openreview.net/forum?id=7pgFL2Dkyyy}.

\bibitem[van~de Ven \& Tolias(2018)van~de Ven and Tolias]{van2018generative}
van~de Ven, G.~M. and Tolias, A.~S.
\newblock Generative replay with feedback connections as a general strategy for
  continual learning.
\newblock \emph{arXiv preprint arXiv:1809.10635}, 2018.

\bibitem[Verma et~al.(2018)Verma, Arora, Mishra, and Rai]{vermageneralized}
Verma, V.~K., Arora, G., Mishra, A., and Rai, P.
\newblock Generalized zero-shot learning via synthesized examples.
\newblock \emph{Proceedings of the IEEE/CVF Conference on Computer Vision and
  Pattern Recognition}, 2018.

\bibitem[Verma et~al.(2020)Verma, Brahma, and Rai]{verma2019meta}
Verma, V.~K., Brahma, D., and Rai, P.
\newblock A meta-learning framework for generalized zero-shot learning.
\newblock \emph{Association for the Advancement of Artificial Intelligence},
  2020.

\bibitem[Verma et~al.(2021)Verma, Mishra, Pandey, Murthy, and
  Rai]{verma2021towards}
Verma, V.~K., Mishra, A., Pandey, A., Murthy, H.~A., and Rai, P.
\newblock Towards zero-shot learning with fewer seen class examples.
\newblock In \emph{Proceedings of the IEEE/CVF Winter Conference on
  Applications of Computer Vision}, pp.\  2241--2251, 2021.

\bibitem[Vitter(1985)]{vitter1985random}
Vitter, J.~S.
\newblock Random sampling with a reservoir.
\newblock \emph{ACM Transactions on Mathematical Software (TOMS)}, pp.\
  37--57, 1985.

\bibitem[Vyas et~al.(2020)Vyas, Venkateswara, and
  Panchanathan]{vyas2020leveraging}
Vyas, M.~R., Venkateswara, H., and Panchanathan, S.
\newblock Leveraging seen and unseen semantic relationships for generative
  zero-shot learning.
\newblock In \emph{European Conference on Computer Vision}, pp.\  70--86.
  Springer, 2020.

\bibitem[Wah et~al.(2011)Wah, Branson, Welinder, Perona, and Belongie]{CUB}
Wah, C., Branson, S., Welinder, P., Perona, P., and Belongie, S.
\newblock The caltech-ucsd birds-200-2011 dataset.
\newblock 2011.

\bibitem[Wei et~al.(2020)Wei, Deng, and Yang]{wei2020lifelong}
Wei, K., Deng, C., and Yang, X.
\newblock Lifelong zero-shot learning.
\newblock \emph{IJCAI}, 2020.

\bibitem[Wu \& He(2018)Wu and He]{wu2018group}
Wu, Y. and He, K.
\newblock Group normalization.
\newblock In \emph{Proceedings of the European conference on computer vision
  (ECCV)}, pp.\  3--19, 2018.

\bibitem[Xian et~al.(2016)Xian, Akata, Sharma, Nguyen, Hein, and
  Schiele]{xian2016latent}
Xian, Y., Akata, Z., Sharma, G., Nguyen, Q., Hein, M., and Schiele, B.
\newblock Latent embeddings for zero-shot classification.
\newblock In \emph{Proceedings of the IEEE conference on computer vision and
  pattern recognition}, pp.\  69--77, 2016.

\bibitem[Xian et~al.(2018{\natexlab{a}})Xian, Lampert, Schiele, and
  Akata]{xian2018zero}
Xian, Y., Lampert, C.~H., Schiele, B., and Akata, Z.
\newblock Zero-shot learning-a comprehensive evaluation of the good, the bad
  and the ugly.
\newblock \emph{IEEE transactions on pattern analysis and machine
  intelligence}, 2018{\natexlab{a}}.

\bibitem[Xian et~al.(2018{\natexlab{b}})Xian, Lorenz, Schiele, and
  Akata]{xian2018feature}
Xian, Y., Lorenz, T., Schiele, B., and Akata, Z.
\newblock Feature generating networks for zero-shot learning.
\newblock In \emph{Proceedings of the IEEE Conference on Computer Vision and
  Pattern Recognition}, 2018{\natexlab{b}}.

\bibitem[Xian et~al.(2019{\natexlab{a}})Xian, Sharma, Schiele, and
  Akata]{f-VAEGAN-D2}
Xian, Y., Sharma, S., Schiele, B., and Akata, Z.
\newblock F-vaegan-d2: A feature generating framework for any-shot learning.
\newblock In \emph{The IEEE Conference on Computer Vision and Pattern
  Recognition (CVPR)}, June 2019{\natexlab{a}}.

\bibitem[Xian et~al.(2019{\natexlab{b}})Xian, Sharma, Schiele, and
  Akata]{xian2019f}
Xian, Y., Sharma, S., Schiele, B., and Akata, Z.
\newblock f-vaegan-d2: A feature generating framework for any-shot learning.
\newblock In \emph{Proceedings of the IEEE Conference on Computer Vision and
  Pattern Recognition}, pp.\  10275--10284, 2019{\natexlab{b}}.

\bibitem[Xu \& Zhu(2018)Xu and Zhu]{xu2018reinforced}
Xu, J. and Zhu, Z.
\newblock Reinforced continual learning.
\newblock In \emph{Advances in Neural Information Processing Systems}, pp.\
  899--908, 2018.

\bibitem[Yoon et~al.(2020)Yoon, Kim, Yang, and Hwang]{yoon2020scalable}
Yoon, J., Kim, S., Yang, E., and Hwang, S.~J.
\newblock {Scalable and Order-robust Continual Learning with Additive Parameter
  Decomposition}.
\newblock \emph{International Conference on Learning Representations},
  abs/1902.09432, 2020.

\bibitem[Yu \& Lee(2019)Yu and Lee]{yu2019zero}
Yu, H. and Lee, B.
\newblock Zero-shot learning via simultaneous generating and learning.
\newblock In \emph{Advances in Neural Information Processing Systems}, pp.\
  46--56, 2019.

\bibitem[Yu et~al.(2020{\natexlab{a}})Yu, Twardowski, Liu, Herranz, Wang,
  Cheng, Jui, and Weijer]{yu2020semantic}
Yu, L., Twardowski, B., Liu, X., Herranz, L., Wang, K., Cheng, Y., Jui, S., and
  Weijer, J. v.~d.
\newblock Semantic drift compensation for class-incremental learning.
\newblock In \emph{Proceedings of the IEEE/CVF Conference on Computer Vision
  and Pattern Recognition}, pp.\  6982--6991, 2020{\natexlab{a}}.

\bibitem[Yu et~al.(2020{\natexlab{b}})Yu, Ji, Han, and Zhang]{yu2020episode}
Yu, Y., Ji, Z., Han, J., and Zhang, Z.
\newblock Episode-based prototype generating network for zero-shot learning.
\newblock In \emph{Proceedings of the IEEE/CVF Conference on Computer Vision
  and Pattern Recognition}, pp.\  14035--14044, 2020{\natexlab{b}}.

\bibitem[Zhang \& Saligrama(2016)Zhang and
  Saligrama]{saligram2016learningJoint}
Zhang, Z. and Saligrama, V.
\newblock Learning joint feature adaptation for zero-shot recognition.
\newblock \emph{arXiv preprint arXiv:1611.07593}, 2016.

\bibitem[Zhu et~al.(2019)Zhu, Xie, Tang, Peng, and Elgammal]{zhu2019semantic}
Zhu, Y., Xie, J., Tang, Z., Peng, X., and Elgammal, A.
\newblock Semantic-guided multi-attention localization for zero-shot learning.
\newblock In \emph{Advances in Neural Information Processing Systems}, pp.\
  14943--14953, 2019.

\end{thebibliography}
\bibliographystyle{icml2021}

\clearpage

\appendix

\section{Dataset Descriptions}
We conduct experiments on five widely used datasets for zero-shot learning. CUB-200~\cite{CUB} is a fine-grain dataset containing 200 classes of birds, and AWA1~\cite{AWA1} and AWA2~\cite{xian2018zero} are datasets containing 50 classes of animal, each represented by an $85$-dimensional attribute. aPY~\cite{aPY} is a diverse dataset containing 32 classes, each associated with a 64-dimensional attribute. SUN~\cite{patterson2012sun} includes $717$ classes, each with only 20 samples; fewer samples and a high number of classes make SUN especially challenging. In the SUN dataset, each class is represented by a 102-dimensional attribute vector. The details of the train/test split are summarized in Table-\ref{tab:dataset}; the same splits are used for the generalized zero-shot Learning (GZSL) setting.

We use the publicly available\footnote{\href{1}{http://datasets.d2.mpi-inf.mpg.de/xian/xlsa17.zip}} pre-processed dataset provided by \citet{xian2018zero}. For each visual domain, we use ResNet-101~\cite{he2016deep} features pre-trained on ImageNet~\cite{imagenet2015}. Features are directly extracted from the pre-trained model without any finetuning. Note that the seen and unseen splits proposed by \cite{xian2018zero} ensure that unseen classes are not present in the ImageNet dataset; otherwise, the pre-trained weights would violates the zero-shot learning setting. 

\section{Reservoir Replay and Task Details for Fixed Continual GZSL}
In the fixed continual GZSL setting, all classes appear during train and test for each class, either as seen or unseen. Initially, a small subset of classes are considered seen, while the rest are considered unseen; which each task, training data for an increasing number of the unseen classes becomes available, making them the new seen classes.
For the AWA1 and AWA2 datasets, which contain $C^F_{\mathrm{AWA}}=50$ classes each, classes are divided into five tasks, with ten classes becoming seen per task. The memory reservoir we use is set to $25\times C^F_{\mathrm{AWA}}$. For SUN, we divide the $C^F_{\mathrm{SUN}}=717$ classes of SUN dataset into 15 tasks: 47 unseen classes becoming seen for the first three tasks, and 48 classes becoming seen for each of the remaining tasks. The memory reservoir for SUN is set to $5\times C^F_{\mathrm{SUN}}$. For the CUB dataset, which contains $C^F_{\mathrm{CUB}}=200$ classes, we divide all classes into 20 tasks, for ten classes becoming seen per task. We set memory reservoir to $10\times C^F_{\mathrm{CUB}}$. Finally, we split the $C^F_{\mathrm{aPY}}=32$ classes of the aPY dataset into four tasks of eight classes each. The memory reservoir for the aPY dataset is set to $25\times ^F_{\mathrm{aPY}}$.

\begin{table}[t]
    \centering
    \begin{tabular}{l|p{3em}|p{3em}|p{4.5em}|p{4em}}
    \hline
         Dataset& {Seen Classes}& Unseen Classes&Attribute Dimension&Total Classes  \\
         \hline
         AWA1& 40&10&85&50\\
         AWA2& 40&10&85&50\\
         CUB& 150&50&312&200\\
         SUN& 645&72&102&717\\
         aPY& 20&12&64&32\\
         \hline
    \end{tabular}
    \caption{Datasets and their seen/unseen class splits for the dynamic GZSL setting.}
    \label{tab:dataset}
\end{table}

\section{Reservoir Replay and Task Details for Dynamic Continual GZSL}
In the dynamic continual GZSL setting, the seen and unseen classes for each dataset are divided into multiple tasks, and with each task, the number of seen and unseen classes grows. Unlike the fixed continual GZSL setting, new seen classes are completely novel classes, not previously unseen ones. We split the AWA1 and AWA2 datasets to contain $C^D_{\mathrm{AWA}}=40$ seen classes and 10 unseen classes, evenly divided into five tasks; thus for each task, we have 8 seen and 2 unseen classes. The reservoir memory for the AWA1 and AWA2 datasets is set to $25\times C^D_{\mathrm{AWA}}$.
The CUB dataset is split to have $C^D_{\mathrm{CUB}} = 150$ seen and 50 unseen classes, divided into 20 tasks; the initial ten tasks are assigned seven seen and two unseen classes each, while the remaining ten tasks have eight seen and three unseen class. The reservoir memory for the CUB dataset is set to $10\times C^D_{\mathrm{CUB}}$. The aPY dataset has $C^D_{\mathrm{aPY}} = 20$ seen and 12 unseen classes divided into four tasks, with each task having five seen classes and three unseen classes. The reservoir memory for the aPY dataset contains $25\times C^D_{\mathrm{aPY}}$. Similarly, the SUN dataset is split into 15 tasks: 43 seen and 4 unseen classes in the first three tasks, with the remaining 12 tasks containing 43 seen and five unseen classes. The reservoir memory for the SUN dataset contains $5\times C^D_{\mathrm{SUN}}$, where the number of seen classes is $C^D_{\mathrm{SUN}}=645$.
We summarize the seen and unseen class splits and attribute dimension for each dataset in Table~\ref{tab:dataset}.

\section{Implementation Details}
For our experiments, we implement self-gating modules $\Phi_a$, $\Phi_s$, and $\Phi_b$ each as a single fully connected layer with ReLU, Sigmoid, and ReLU activation functions. The hidden dimension of each neural network module is set to $2048$, on top of which scaled class normalization is applied. The self-gating output of a given attribute is then sent to another one-layer neural network of dimension $2048\rightarrow 2048$, and scaled class normalization is applied again. This output is then considered the projected visual features. In this visual space, we measure similarity by cosine distance. 

For each task of all datasets, the model is trained for 200 epochs. For the inner training loop we use an Adam optimizer~\cite{kingma2014adam} with a constant learning rate $0.0001$. For the meta update, we use an Adam optimizer with an initial learning rate of $0.001$, decreasing with each epoch at a rate of $1- e/(E-1)$, where $e$ and $E$ are the current epoch and total number of epochs, respectively. We use the same hyperparameter settings for all dataset, finding our model to be stable, with applicability to a diverse set of datasets.

\end{document}